\documentclass[runningheads]{llncs}

\pdfoutput=1
\usepackage{eccv}

\usepackage{eccvabbrv}

\usepackage{graphicx}
\usepackage{booktabs}

\usepackage[accsupp]{axessibility}  
\usepackage{pifont}
\newcommand{\xmark}{\ding{55}} 
\usepackage{colortbl}
\usepackage{multirow}
\usepackage{makecell}
\usepackage{caption}
\usepackage{xspace}
\usepackage{marvosym}
\usepackage{fontawesome5}
\usepackage{graphicx}
\usepackage{wrapfig}

\usepackage[colorlinks=true, linkcolor=red, urlcolor=blue, citecolor=green]{hyperref}
\usepackage{orcidlink}

\begin{document}

\title{Out of Sight but Not Out of Mind: Hybrid Memory for Dynamic Video World Models} 

\titlerunning{Hybrid Memory for Dynamic Video World Models}

\author{Kaijin Chen\inst{1} \and
Dingkang Liang\inst{1} \and
Xin Zhou\inst{1} \and Yikang Ding\inst{2} \and Xiaoqiang Liu\inst{2} \and Pengfei Wan\inst{2}  \and Xiang Bai\inst{1}}

\authorrunning{K. Chen et al.}

\institute{Huazhong University of Science and Technology \and
Kling Team, Kuaishou Technology \\
\email{\{kjchen, dkliang\}@hust.edu.cn} \\
Project Page: \href{https://kj-chen666.github.io/Hybrid-Memory-in-Video-World-Models/}{\textcolor{blue}{Hybrid-Memory-in-Video-World-Models}} 
}

\maketitle
\begingroup
\renewcommand\thefootnote{}
\footnotetext{Work done during an internship at Kling Team, Kuaishou Technology.}
\endgroup

\begin{center}
    \includegraphics[width=\linewidth]{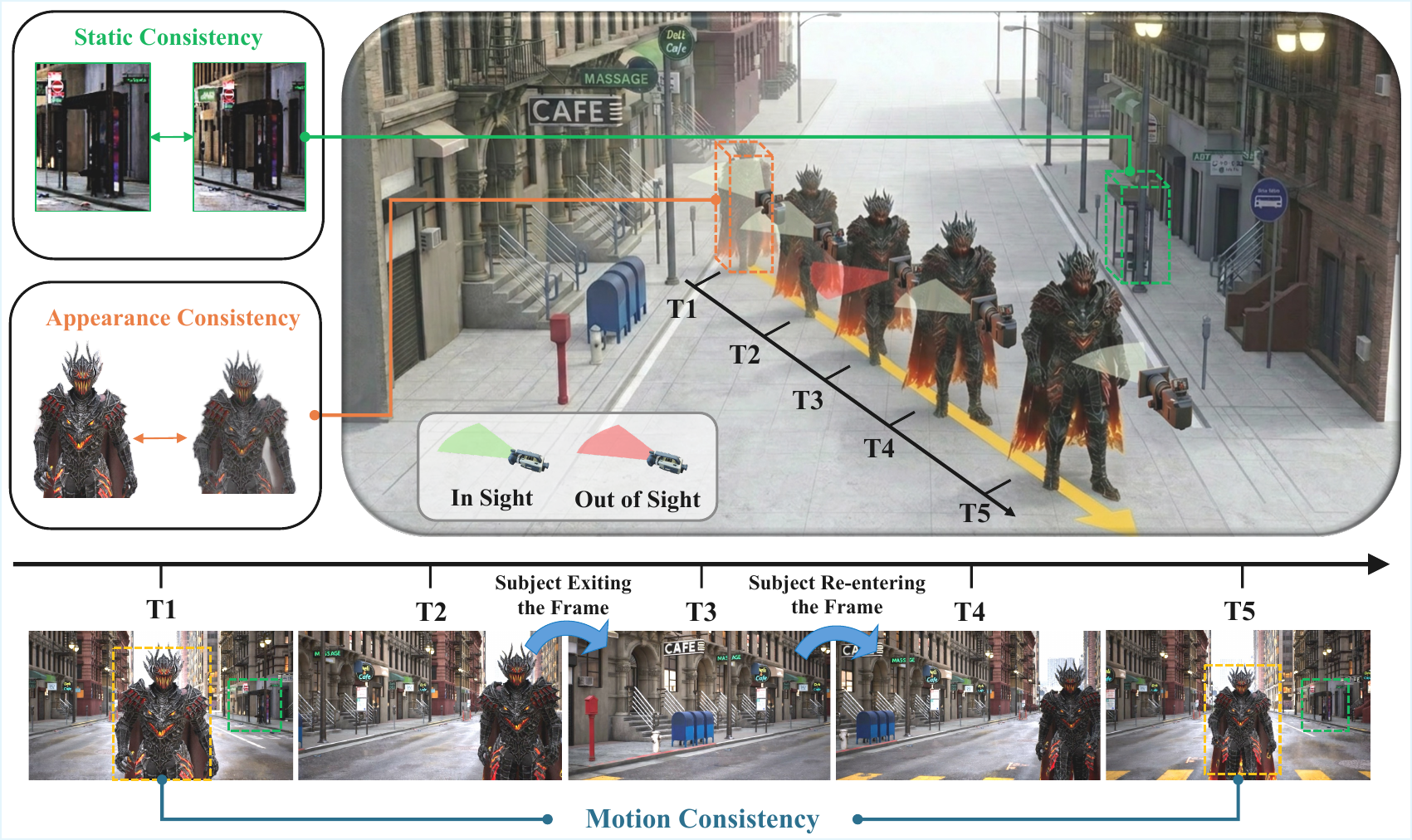}
    \vspace{-17pt} 
    \captionof{figure}{Hybrid Memory demands the model to maintain static consistency in backgrounds, while simultaneously preserving the motion and appearance consistency of dynamic subjects during out-of-view intervals.}
    \label{fig:intro}
\end{center}
\vspace{-10pt}

\begin{abstract}
  Video world models have shown immense potential in simulating the physical world, yet existing memory mechanisms primarily treat environments as static canvases. When dynamic subjects hide out of sight and later re-emerge, current methods often struggle, leading to frozen, distorted, or vanishing subjects. To address this, we introduce \textbf{Hybrid Memory}, a novel paradigm requiring models to simultaneously act as precise archivists for static backgrounds and vigilant trackers for dynamic subjects, ensuring motion continuity during out-of-view intervals. To facilitate research in this direction, we construct \textbf{HM-World}, the first large-scale video dataset dedicated to hybrid memory. It features 59K high-fidelity clips with decoupled camera and subject trajectories, encompassing 17 diverse scenes, 49 distinct subjects, and meticulously designed exit-entry events to rigorously evaluate hybrid coherence. Furthermore, we propose \textbf{HyDRA}, a specialized memory architecture that compresses memory into tokens and utilizes a spatiotemporal relevance-driven retrieval mechanism. By selectively attending to relevant motion cues, HyDRA effectively preserves the identity and motion of hidden subjects. Extensive experiments on HM-World demonstrate that our method significantly outperforms state-of-the-art approaches in both dynamic subject consistency and overall generation quality. Code is publicly
  available at \url{https://github.com/H-EmbodVis/HyDRA}.
\end{abstract}

\section{Introduction}
\label{sec:intro}

World Models \cite{cui2025emu3, he2025matrix, mao2025yume, ye2025yan} have recently garnered significant research attention for their ability to generate high-fidelity environments that align with the real world. These models have demonstrated immense potential across diverse downstream domains, including autonomous driving\cite{gao2024vista, zhou2025hermes, liang2026UniFuture} and embodied intelligence\cite{wang2025embodiedgen, jiang2025enerverse}. The latest advancements in video generation \cite{wan2025wan, kong2024hunyuanvideo, hong2022cogvideo} further validate the feasibility of modeling the physical world. Crucially, memory mechanisms have emerged as a critical frontier in advancing world models, as memory capacity dictates the spatial and temporal consistency of generated content. Specifically, it is the cognitive anchor that allows the model to retain historical context during viewpoint shifts or long-term extrapolation. Without robust memory, a simulated world quickly unravels into disconnected, chaotic frames.

While recent studies \cite{yu2025context, li2025vmem, xiao2025worldmem, huang2025memory, wu2025pack} have enhanced the memory capacity through advanced retrieval retrieval\cite{yu2025context, xiao2025worldmem, li2025vmem} and compression\cite{wu2025pack} techniques, they share a common blind spot: treating the world as a static canvas. They excel at memorizing and reconstructing motionless environments, but the physical world is a bustling, dynamic stage populated by subjects (e.g., walking pedestrians, running animals) governed by their independent motion logic. When dynamic subjects hide outside the camera's field of view, these models lose track of them, often rendering the returning subjects as frozen statues, distorted phantoms, or simply letting them vanish into the air. To bridge this gap, we introduce a novel memory paradigm: \textbf{Hybrid Memory}, which requires the model to simultaneously perform precise memorization and viewpoint reconstruction of static backgrounds, while continuously seeking and predicting the motion of dynamic subjects.  As illustrated in Fig.~\ref{fig:intro}, when a subject hides out of view, the model must not only remember its appearance but also mentally predict its unseen trajectory, ensuring both visual coherence and motion consistency when they re-enter the frame.

To investigate and validate this new hybrid memory paradigm, constructing a specialized dataset and designing corresponding memory mechanisms are imperative. In this work, we introduce \textbf{HM-World}, the first large-scale video dataset purpose-built to train and evaluate \textbf{H}ybrid \textbf{M}emory capabilities. HM-World possesses two core properties: 1) meticulously designed shots with dynamic subjects exiting and entering the frame, and 2) highly diverse scenarios, subjects, and motion patterns. Comprising \textbf{59K} video clips, the dataset deliberately decouples camera trajectories from subject movements, creating countless natural instances where subjects slip into the unseen margins before re-emerging.  Furthermore, HM-World exhibits exceptional diversity, encompassing 17 distinctively styled scenes, 49 different subjects (including humans of various appearances and multiple animal species), 10 motion paths for subjects, and 28 types of camera trajectories.

Based on the proposed dataset HM-World, we evaluate existing methods and observe that they tend to either immobilize moving objects or distort dynamic content, lacking the hybrid memory capacity to track unseen motion. To equip models with this capacity, we propose \textbf{HyDRA} (\textbf{Hy}brid \textbf{D}ynamic \textbf{R}etrieval \textbf{A}ttention), a memory approach designed to seek the hidden subjects and preserve dynamic consistency. HyDRA employs a Memory Tokenizer that compresses memory latents into tokens with richer information. When a subject is poised to re-enter the frame, HyDRA utilizes a spatiotemporal relevance-driven retrieval mechanism to actively scan these tokens, pulling the most crucial motion and appearance cues into the current denoising process. This allows the model to effectively rediscover the hidden subject, seamlessly picking up its trajectory where it left off. Extensive experiments on HM-World demonstrate that HyDRA significantly outperforms state-of-the-art approaches in preserving dynamic subject consistency and overall generation quality. Ablation studies further verify the robustness of our design. We hope our dataset and method can offer a fresh perspective for the community.

Our main contributions can be summarized as follows: \textbf{1)} We identify the limitations of existing static-centric memory mechanisms and propose \textbf{Hybrid Memory}, a novel paradigm that requires models to simultaneously maintain spatial consistency for static backgrounds, and motion continuity for dynamic subjects, especially during out-of-view intervals. \textbf{2)} We introduce \textbf{HM-World}, the first large-scale video dataset dedicated to hybrid memory research. Featuring 59K clips with diverse scenes, subjects, and motion patterns,  it provides a rigorous benchmark for evaluating  spatiotemporal coherence in complex, dynamic environments. \textbf{3)} We propose \textbf{HyDRA}, a specialized memory architecture that utilizes a spatiotemporal relevance-driven retrieval mechanism with memory tokens. By attending to relevant motion cues, HyDRA effectively seeks and rediscovers hidden subjects and preserves its identity and motion, significantly outperforming existing state-of-the-art methods.

\vspace{-5pt}

\section{Related Works}

\subsection{Video World Models}
\vspace{-3pt}
Recent advances in video generation models \cite{wan2025wan, kong2024hunyuanvideo, hong2022cogvideo, zheng2024open, huang2025selfforcing, zhou2025less} have demonstrated their potential in modeling the real world and synthesizing high-fidelity clips, increasingly serving as the foundation for world models. Building on this progress, multiple video world models have been introduced \cite{che2024gamegen, mao2025yume, he2025matrix, sun2025worldplay, hong2025relic, li2025hunyuan, bar2025navigation}. GameGen-X\cite{che2024gamegen} explores interactive video world models within game-like environments. Yume \cite{mao2025yume} further increases the length of generated videos through autoregressive generation. Matrix-Game 2 \cite{he2025matrix} constructs a large-scale dataset based on GTA-V and Unreal Engine 5 \cite{UnrealEngine5} and incorporates autoregressive denoising \cite{huang2025selfforcing} to achieve controllability and visual quality comparable to video games. RELIC \cite{hong2025relic} focuses on static scene consistency and distills long‑video generation with replayed back-propagation, enabling stable, long‑duration generation. Worldplay \cite{sun2025worldplay} leverages large-scale, high‑quality data and context forcing technique to deliver both exceptional visual quality and consistency while supporting real‑time generation.

Despite significant progress, video world models continue to confront several challenges, with generation consistency being a prominent one. Current models still struggle to maintain both static and dynamic consistency across generated sequences. This issue is particularly pronounced during long-duration generation and under camera motion, where models frequently lose track of previously generated content or contextual input, leading to inconsistent outputs. Our work aims to tackle this challenge from the perspective of hybrid memory, enabling spatiotemporally consistent generation.

% \vspace{-18pt}
\subsection{Memory in Video Generation}
Existing memory approaches primarily focus on processing the context and optimizing the interaction and propagation of contextual information during the generation process. Vmem\cite{li2025vmem} employs a 3D surfel-indexed memory structure to retrieve context, while Context-as-Memory \cite{yu2025context} adopts Field-of-View (FOV) overlap. Worldmem\cite{xiao2025worldmem} combines FOV-based retrieval for an external memory bank with Diffusion Forcing \cite{chen2024diffusion} on Minecraft data. Memory Forcing\cite{huang2025memory} further incorporates temporal memory to balance exploration and consistency. Similarly, WorldPlay\cite{sun2025worldplay} enhances long-term generation consistency through a context-forcing approach. Inspired by FramePack \cite{zhang2025packing}, MemoryPack\cite{wu2025pack} introduces an updatable semantic pack throughout the generation process, retaining semantically relevant memory. In parallel, RELIC\cite{hong2025relic} applies uniform spatial down-sampling to compress context memory.

Existing studies have achieved notable results. However, most of these methods are designed for static scenes\cite{yu2025context, li2025vmem, hong2025relic} or relatively simple dynamic environments\cite{xiao2025worldmem, huang2025memory, wu2025pack}, and have not been specifically optimized for complex dynamic scenes involving moving subjects and dynamic elements. Although Genie 3 \cite{genie3} demonstrates remarkable dynamic consistency, it is a closed-source model with technical details remaining undisclosed. This research gap persists in both dataset construction and method design. To address this, our work focuses on hybrid memory in complex dynamic scenes, tackling the challenge from both methodological and dataset perspectives.

\section{HM-World: Dataset}
To address the research gap in hybrid memory, we conduct an in-depth analysis of its definition and inherent challenges for current video world models in Sec.\ref{sec:bench1}. Building upon this analysis, we introduce \textbf{HM-World}, a large-scale dataset constructed for \textbf{H}ybrid \textbf{M}emory in Video \textbf{World} Models, and detail its characteristics in Sec.~\ref{sec:bench2}.

% \vspace{-5pt}
\begin{figure}[t]
	\begin{center}
		\includegraphics[width=\linewidth]{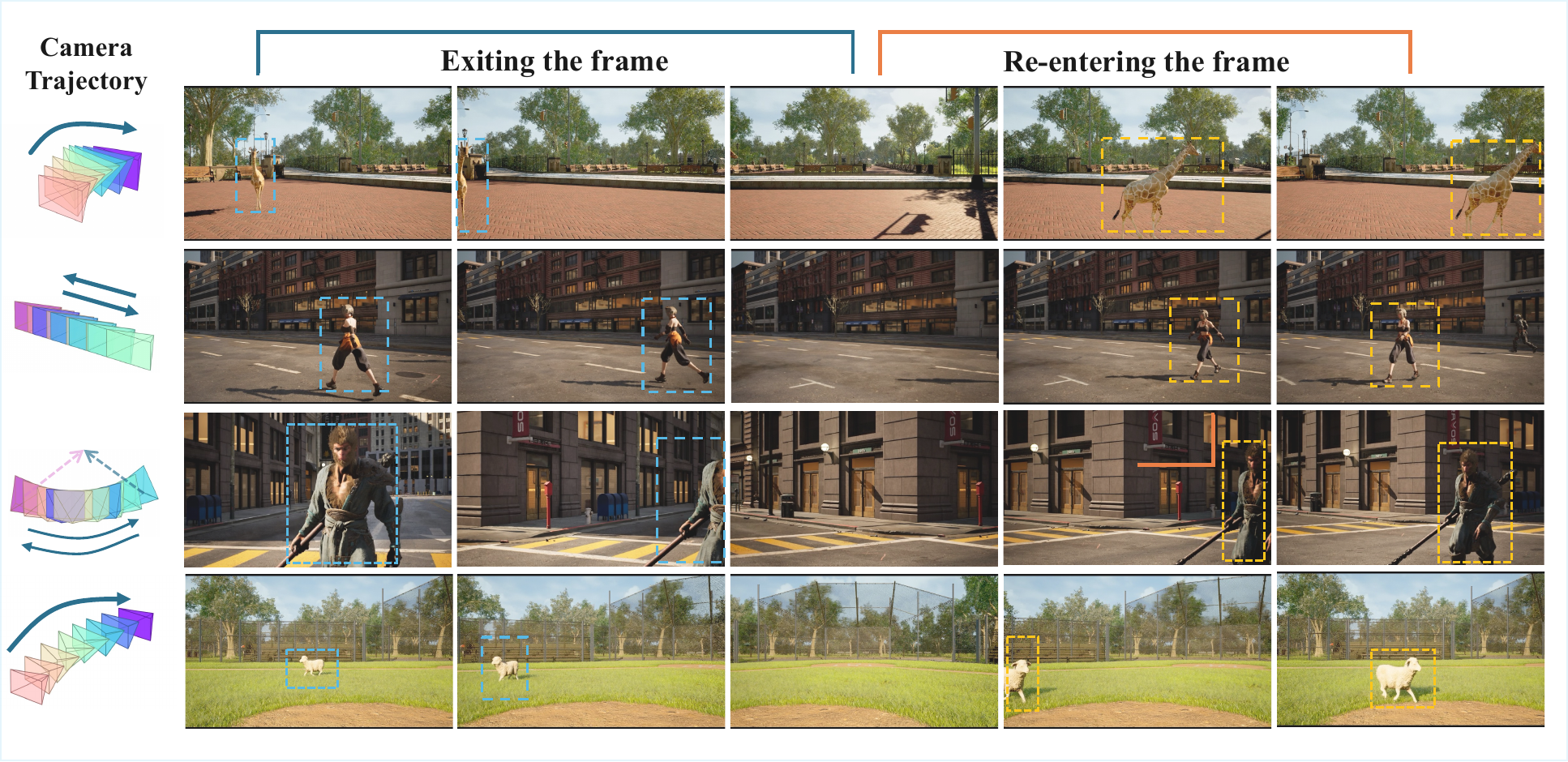}
	\end{center}
	\vspace{-15pt}
	\caption{Instances of exit-entry camera motion.
 }
 \vspace{-12pt}
	\label{fig:camera motions}
\end{figure}

\subsection{Hybrid Memory}

% \vspace{-15pt}
\label{sec:bench1}
Memory refers to the model's ability to retain information from inputs or generated content, ensuring consistency throughout the generation process. Static memory ensures the consistency of immobile elements (e.g., buildings, roads), and is typically evaluated by assessing whether a scene looks identical when the camera returns to a previous pose\cite{yu2025context}. Hybrid memory, however, demands a far more sophisticated cognitive leap. It requires the model to simultaneously anchor the static background while tracking the dynamic subjects (e.g., pedestrians, running dogs). As illustrated in Fig.~\ref{fig:camera motions}, when a subject exits and re-enters the frame, hybrid memory dictates that it must not only retain its original visual identity but also reappear at a plausible location with a consistent motion state.

Achieving hybrid memory is challenging for several reasons: 1) \textbf{Need for spatiotemporal decoupling}. Unlike static memory, which merely maps camera poses to a fixed 3D space, hybrid memory forces the model to independently untangle the camera's ego-motion from the subject's independent trajectory. 2) \textbf{Out-of-view extrapolation}. Once a subject steps off-stage, the model loses direct visual evidence and must implicitly simulate the subject's movement in the latent space. 3) \textbf{Feature entanglement}. In standard diffusion latents, static background features and subject features are heavily coupled. Retrieving historical context without isolating the dynamic cues often causes the subjects to freeze into the background or distort unnaturally.

To conquer these complex dynamics and bridge the research gap, a dedicated testing ground is essential. As natural videos with perfectly captured, unoccluded exit-and-re-entry events are remarkably scarce, we constructed HM-World, a dataset explicitly tailored for hybrid memory.
\vspace{-5pt}
\subsection{Dataset Characteristics}
\label{sec:bench2}

\begin{figure}[t]
	\begin{center}
		\includegraphics[width=\linewidth]{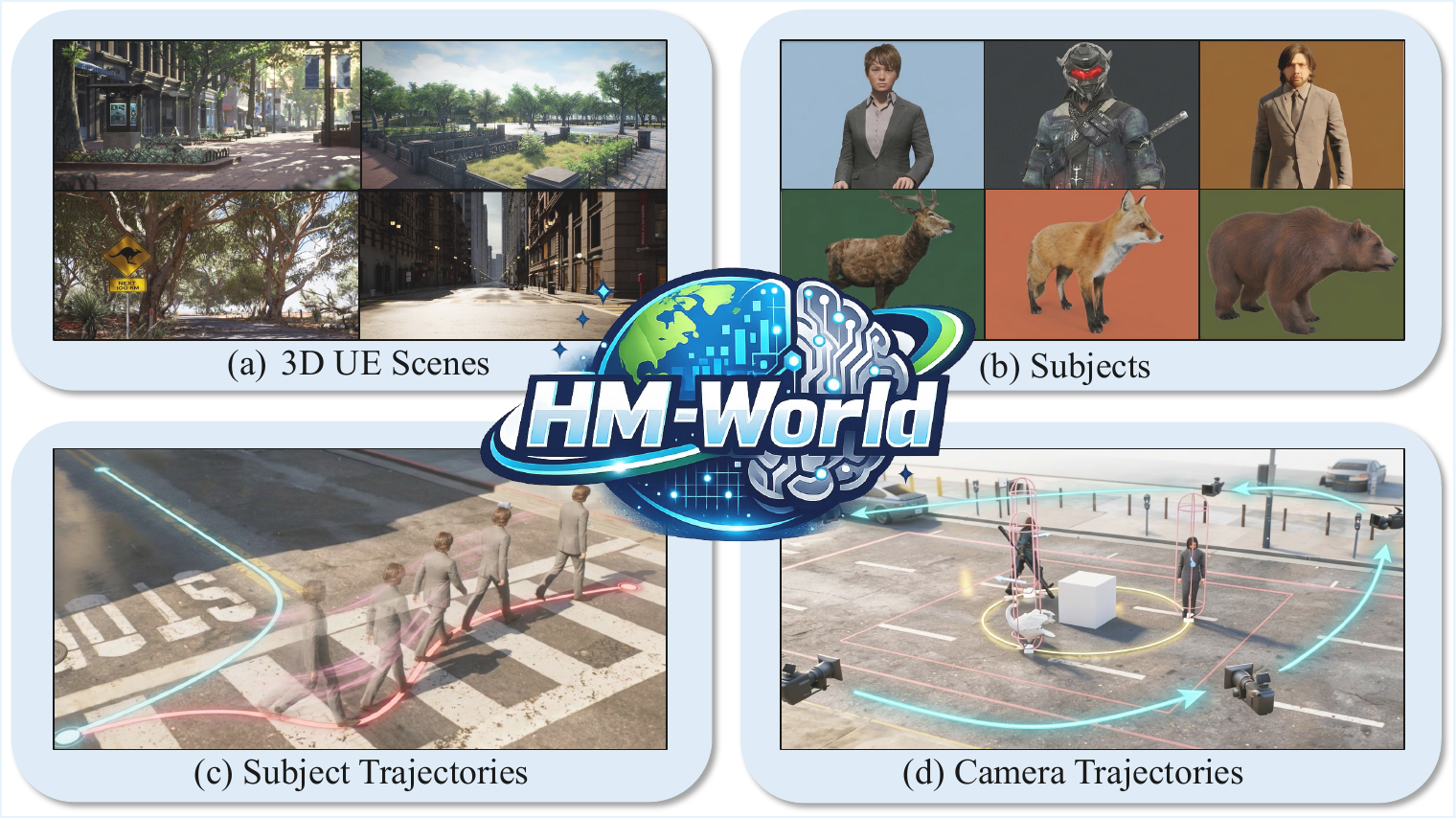}
	\end{center}
	\vspace{-15pt}
	\caption{Construction Procedure of HM-World. We combine (a) 3D scenes, (b) subjects, (c) subject trajectories, and (d) camera trajectories to render data containing dynamics in Unreal Engine 5.
 }
 \vspace{-15pt}
	\label{fig:dataset}
\end{figure}

Since videos with exit-entry events are rarely found on the Internet, we construct the dataset by implementing a data rendering pipeline within Unreal Engine 5 \cite{UnrealEngine5}. As depicted in Fig.~\ref{fig:dataset}, our data generation process is structured along four dimensions: scenes, subjects, subject trajectories, and camera trajectories. We first collect 17 stylistically diverse scenes to serve as the environmental background. Then, 49 distinct subjects, encompassing people of varied appearances and animals of multiple species, are combined into groups of 1 to 3. Each combination is procedurally placed within a scene. Furthermore, each subject is associated with its own motion animation and follows a randomly selected trajectory from a set of 10 predefined paths. 

To guarantee a rich density of exit-entry events, we meticulously designed the camera motions. Moving beyond simple unidirectional tracking, our camera trajectories incorporate deliberate back‑and‑forth camera motions, as illustrated in Fig.~\ref{fig:camera motions}, to actively induce hide-and-reappear dynamics. For instance, a leftward pan followed by a rightward pan typically causes a captured subject to leave and re-enter the frame. Following this principle, we designed 28 distinct camera trajectories. Additionally, each camera movement is assigned multiple initial positions, further enhancing the diversity of camera motion sequences. 

\begin{table*}[t!]
\centering
\scriptsize
\renewcommand{\tabcolsep}{1.2mm} 
\caption{The comparison between existing datasets and HM-World dataset. "Dynamic Subject" means including moving subjects, "Exit-Enter" refers to containing exit-entry events in clips, and "Subject Pose" denotes including annotated 3D poses of subjects.}
\vspace{-10pt}
\begin{tabular}{@{}llccccc@{}} 
\toprule
Dataset & Reference & \begin{tabular}[c]{@{}c@{}}Dynamic\\ Subject\end{tabular} & \begin{tabular}[c]{@{}c@{}}Subject\\ Exit-Enter\end{tabular} & \begin{tabular}[c]{@{}c@{}}Subject\\ Pose\end{tabular} & \begin{tabular}[c]{@{}c@{}}Camera\\ Movable\end{tabular} & \begin{tabular}[c]{@{}c@{}}Total\\ Num.\end{tabular} \\
\midrule
WorldScore~\cite{duan2025worldscore} & ICCV 25   & \checkmark & \xmark & \xmark & \checkmark & 3K \\
Context-As-Memory~\cite{yu2025context} & SIGGRAPH Asia 25  &\xmark   & \xmark & \xmark &\checkmark&  10K\\
Multi-Cam Video~\cite{bai2025recammaster} & ICCV 25& \checkmark & \xmark  & \xmark & \checkmark&  136K\\
360\textdegree-Motion~\cite{xiao20243dtrajmaster}    & ICLR 25 & \checkmark & \xmark  & \checkmark & \xmark & 5.4K\\
\midrule
\textbf{HM-World (ours)}& - & \checkmark  &\checkmark &  \checkmark  &\checkmark& 59K\\
\bottomrule
\end{tabular}
\vspace{-15pt}
\label{tab:static}
\end{table*}

After procedurally combining elements from all four dimensions and filtering clips that lack exit‑entry events, we obtain a final collection of \textbf{59,225} high‑fidelity video clips. Every sample is comprehensively annotated with the rendered video, a descriptive caption generated by MiniCPM-V \cite{yao2024minicpm}, corresponding camera poses, per‑frame positions of all subjects, and precise timestamps marking each subject’s exit from and entry into the frame. Tab.~\ref{tab:static} highlights the comparison between HM-World and existing datasets. Specifically, the Context-as-Memory dataset only contains static scenes. WorldScore includes numerous real-world scenes with certain dynamic elements, but its scale is limited to only 3K. Multi-Cam Video features dynamic subjects, but they only perform actions in place. 360 \textdegree-Motion contains moving subjects, but the camera remains static, and the subjects are always within the field of view. In contrast, our HM-World not only features rich, dynamic subjects and complex camera trajectories, but also includes specific in-and-out-of-frame events for hybrid memory.

\section{Hybrid Dynamic Retrieval Attention}
\label{sec:method}

Given a sequence of context frames $X_{ctx} \in \mathbb{R}^{N \times C \times H \times W}$ and a full sequence of camera trajectory $P= \{P_{ctx}, P_{tgt}\}$ spanning both historical and future timestamps, our goal is to predict the target frames $X_{tgt}\in \mathbb{R}^{M \times C \times H \times W} $. Unlike static scene generation, the context frames $X_{ctx}$ feature dynamic subjects governed by their independent motion. As the camera viewpoint shifts according to $P_{tgt}$ (e.g., panning or rotation), these subjects frequently hide and re-enter the camera's field of view. To synthesize high-fidelity future frames $X_{tgt}$, the model must preserve the static background while seeking the moving subjects to maintain their appearance and motion consistency. To achieve this, we introduce \textbf{HyDRA} (\textbf{Hy}brid \textbf{D}ynamic \textbf{R}etrieval \textbf{A}ttention), a memory method designed to decouple and preserve consistency of dynamic subjects.

\subsection{Base Architecture and Camera Injection}

\begin{wrapfigure}{R}{0.5\textwidth}
  \vspace{-25pt} 
  \centering
  \includegraphics[width=\linewidth]{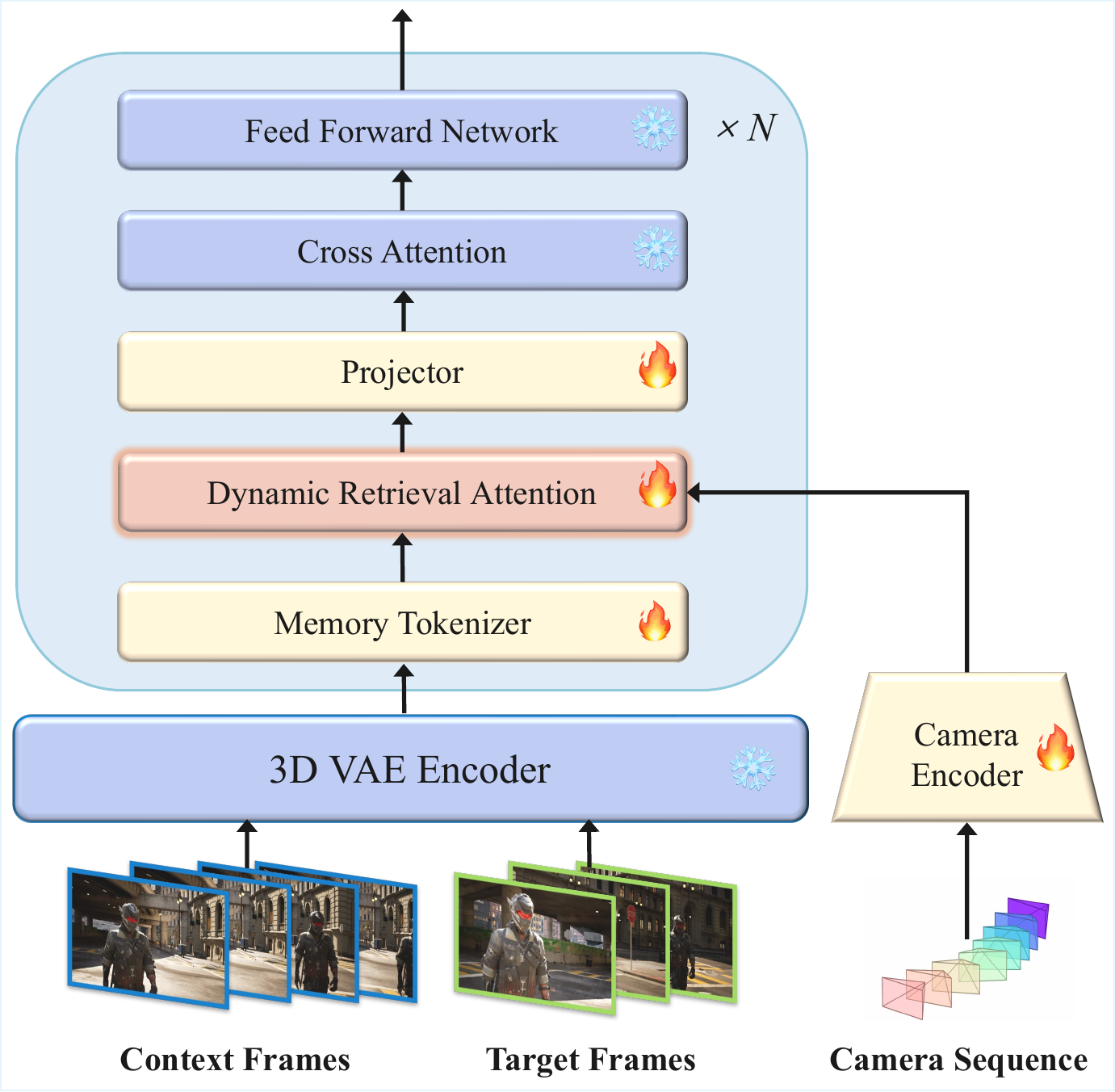} 
  \caption{Model architecture.}
  \label{fig:pipeline}
  \vspace{-25pt} 
\end{wrapfigure}

\textbf{Overall Architecture}. As depicted in Fig.~\ref{fig:pipeline}, our approach is built upon a full-sequence video diffusion model, comprising a causal 3D VAE \cite{kingma2013auto} and a Diffusion Transformer (DiT) \cite{peebles2023scalable}. Each DiT block integrates dynamic retrieval attention, a projector, cross-attention, and a feedforward network (FFN). The diffusion timestep is encoded via a Multi-Layer Perceptron (MLP) to modulate the DiT blocks. The model follows Flow Matching \cite{lipman2022flow}. Given a sequence of video frames $x$, the 3D VAE encodes it into video latent $z_{0} \in \mathbb{R}^{C \times f\times h \times w}$, compressing both temporal and spatial dimensions. During the training phase, the noised latent $z_{t}$ at timestep $t$ is obtained through linear interpolation between  $z_{0}$ and Gaussian noise $z_{1} \sim \mathcal{N}(0, I)$. The model $u$ learns to predict the ground-truth velocity $v_t = z_0 - z_1$ at timestep $t\in [0,1]$, with the loss function defined as:
\begin{equation}
\mathcal{L}_{\theta}=\mathbb{E}_{z_0,z_1,t}||u(z_t,t;\theta)-v_t||^2,
\end{equation}
where $\theta$ represents the model parameters. During the inference phase, randomly sampled Gaussian noise is progressively denoised to yield a clean latent, which is then decoded by the 3D VAE Decoder to reconstruct the video sequence.

\textbf{Camera Injection}. To enable precise spatial control of generated content, we inject camera trajectories into the model as an explicit condition. Suppose the camera pose sequence of length $f$ is denoted as $P = \{ (R_i, t_i) \}_{i=1}^f$, where $R_i \in \mathbb{R}^{3 \times 3}$ and $t_i \in \mathbb{R}^{3}$ represent the rotation matrix and the translation vector for the $i$-th frame, respectively. We flatten and concatenate these parameters to form a unified camera condition $c_{cam} \in \mathbb{R}^{f \times 12}$. Following ReCamMaster\cite{bai2025recammaster}, we employ a camera encoder $\mathcal{E}_{cam}(\cdot)$, implemented as a MLP layer to encode $c_{cam}$.  The encoded camera features are then broadcast spatially and added element-wise to the latent features. Formally, let $H_{in}$ be the sequence features fed into the DiT blocks, the camera-injected feature $H_{out}$ is formulated as:
\begin{equation}
    H_{out} = H_{in} + \mathcal{E}_{cam}(c_{cam}),
\end{equation}
where $\mathcal{E}_{cam}(c_{cam})$ is projected to match the exact channel dimension of $H_{in}$. 

\subsection{Memory Tokenization for Retrieval}
\label{sec:tokenization}
In our framework, the encoded memory latents, denoted as $Z_{mem}$, serve as the primary representation of memory. A naive approach to memory utilization would involve injecting the entire $Z_{mem}$ into the generation process. However, this not only incurs computational overhead but also floods the model with irrelevant noise. Such noise can easily mislead the model's reasoning pathways, ultimately resulting in spatially and temporally inconsistent generation. Therefore, a retrieval mechanism is essential to filter the memory and accurately recall the hidden subject outside the current frame.

Nevertheless, performing retrieval directly on the latent representation could be sub-optimal. Under our proposed hybrid memory paradigm, the task involves highly dynamic subjects and complex spatial relationships driven by camera movements. Direct retrieval from raw, uncoupled latents can lack the expressiveness needed to fully capture the underlying motion of dynamic subjects and the associated camera transformations, potentially undermining spatiotemporal consistency in the generated content.

To overcome this limitation, we introduce a 3D-convolution-based memory tokenizer,  designed to process both spatial and temporal dimensions simultaneously. We argue that facilitating spatiotemporal interaction on the latents yields memory tokens with much deeper, motion-aware representations. This enriched representation is crucial for optimizing the retrieval process and ensuring consistent generation, which is validated by our extensive empirical experiments. 

Specifically, the Memory Tokenizer $\mathcal{T}_{mem}$ processed the latents $Z_{mem}$ into compact memory tokens $M$. By employing 3D convolutions, the tokenizer expands the spatiotemporal receptive field to capture long-duration motion information. Formally, this transformation is defined as:
\begin{equation}
    M = \mathcal{T}_{mem}(Z_{mem}), \quad M \in \mathbb{R}^{C' \times f'_{mem} \times h \times w},
\end{equation}
where $f'_{mem}$ represents the temporal dimension, and $h\times w$ denotes the downsampled spatial resolution. By compressing the raw latents into dense, spatiotemporally-aware memory tokens $M$, the model effectively filters out irrelevant context while preserving the essential motion and appearance cues. These refined tokens $M$ then serve as the foundation for our dynamic retrieval attention module, which will be detailed in the following section.

\begin{figure}[t]
	\begin{center}
		\includegraphics[width=\linewidth]{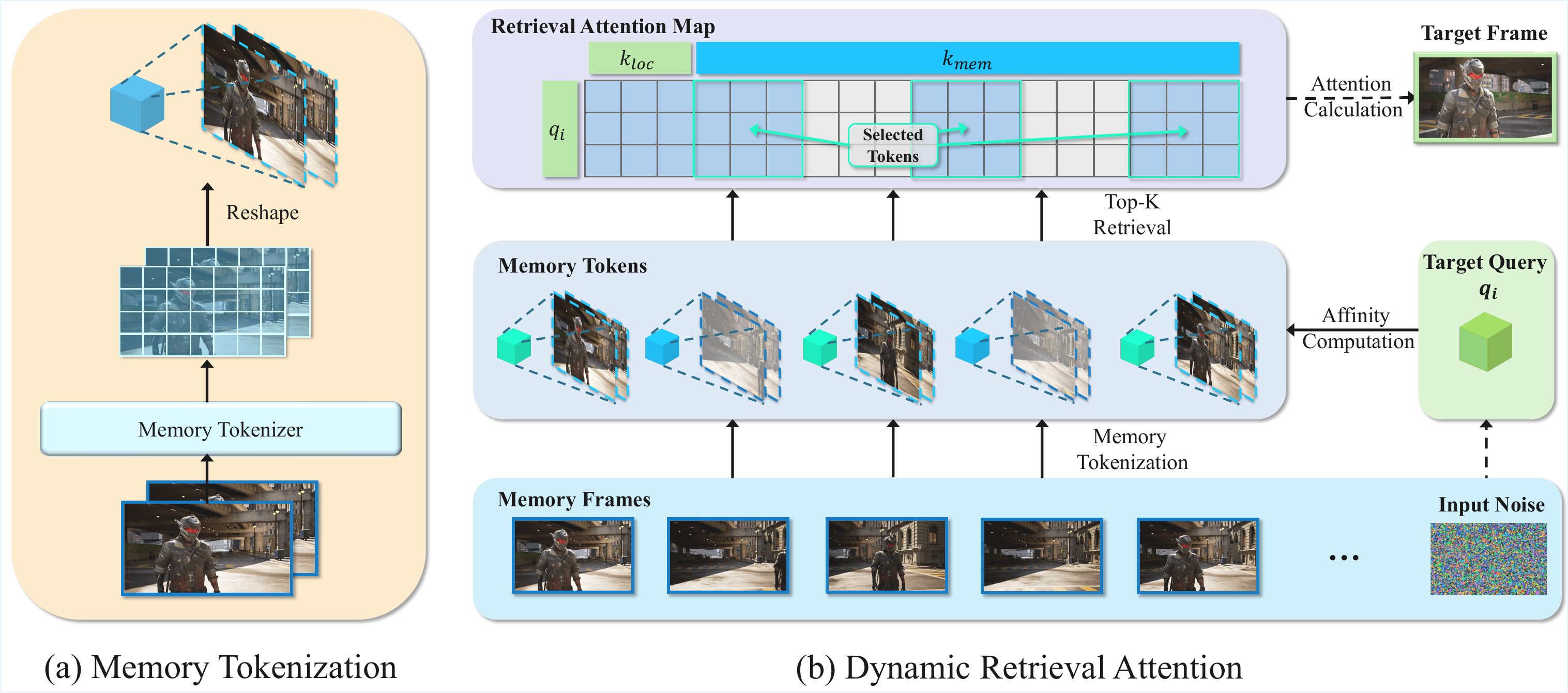}
	\end{center}
	\vspace{-20pt}
	\caption{\textbf{Overview of HyDRA}.  (a) Memory Tokenization Module. (b) Dynamic retrieval attention computes relevance between the target query and memory tokens to retrieve the top-k relevant tokens, enabling the model to recall associated hybrid memory.}
 
 \vspace{-15pt}
	\label{fig:method}
\end{figure}

\subsection{Dynamic Retrieval Attention}
As discussed in Sec.~\ref{sec:tokenization}, indiscriminately injecting all historical context degrades video consistency and inflates computational cost. To tackle this, a retrieval mechanism is imperative for optimizing the information flow. Building upon the principles of attention \cite{vaswani2017attention}, we propose \textbf{Dynamic Retrieval Attention}, a spatiotemporal-informed retrieval method and memory mechanism that directly replaces the standard 3D self-attention layers within the base model.

Given the denoising target latents $Z_{tgt}\in \mathbb{R}^{C' \times f_{tgt} \times H' \times W'}$ and the memory tokens $M \in \mathbb{R}^{C' \times f'_{mem} \times h \times w}$, we first project them into their respective Query, Key, and Value. Concretely, the target latents are projected into queries $Q$, while the memory tokens are projected into keys $K_{mem}$ and values $V_{mem}$.

To perform dynamic retrieval, we process the query set $q_{i}$ corresponding to each target latent $i \in \{1, \dots, f_{tgt}\}$ sequentially. Because $q_{i}$ and $K_{mem}$ operate at different spatial resolutions, we first apply spatial pooling to downsample $q_{i}$ into $\tilde{q}_i \in \mathbb{R}^{C' \times h \times w}$, aligning it with the memory tokens. We then compute a spatiotemporal affinity metric between the downsampled query $\tilde{q}_i$ and each temporal slice of the memory key $k_{mem,j}$ (where $j \in \{1, \dots, f'_{mem}\}$). Since they share the same spatial resolution and channel dimension, the affinity $S_{i,j}$ is calculated by taking the element-wise product across the spatial dimensions:
\begin{equation}
    S_{i,j} = \frac{1}{\sqrt{d}} \sum_{y=1}^{h} \sum_{x=1}^{w} \langle \tilde{q}_{i}(x,y), k_{mem, j}(x,y) \rangle,
\end{equation}
where $\langle \cdot, \cdot \rangle$ denotes the channel-wise inner product, and $d$ is the channel dimension for scaling.

The affinity metric effectively quantifies the spatiotemporal correspondence between the current target latent and the memory token. Based on these affinities, we employ a Top-K selection strategy to filter the memory tokens, isolating the subset of memory that exhibits the strongest correlation with $q_{i}$:
\begin{equation}
    \mathcal{I}_i = \text{TopK}(S_{i}, K), \quad K_{sel} = \{ k_{mem, j} \mid j \in \mathcal{I}_i \}, \quad V_{sel} = \{ v_{mem, j} \mid j \in \mathcal{I}_i \},
\end{equation}
where $\mathcal{I}_i$ represents the indices of the $K$ most relevant memory tokens. 

While retrieving historical memory is crucial for long-term consistency, maintaining local denoising stability is equally important. To preserve the structural integrity of the original self-attention, we forcefully include the queries' own local temporal window into the attention computation. Let $K_{loc}$ and $V_{loc}$ denote the keys and values derived from the adjacent latents within a local window $\mathcal{W}_i$ centered around frame $i$ in the target sequence. We first flatten these local features and the retrieved memory features, then concatenate them to form the final keys $K'_i = [K_{sel}, K_{loc}]$ and values $V'_i = [V_{sel}, V_{loc}]$.

Finally, after flattening the query $q_{i}$, the dynamic retrieval attention for the $i$-th latent is computed using the standard attention formulation: 
\begin{equation}
    \text{Attention}(q_i, K'_i, V'_i) = \text{Softmax}\left( \frac{q_i (K'_i)^T}{\sqrt{d}} \right) V'_i.
\end{equation}
By iterating this process across all queries in the denoising sequence, the model selectively attends to the most pertinent motion and appearance cues of the out-of-sight subjects. Extensive experiments validate that this method successfully tracks hidden subjects,  preserves spatiotemporal consistency, and substantially decreases the computational burden.

\section{Experiments}

\subsection{Experiment Setup}
\textbf{Implement Details}.
We build our method on Wan2.1-T2V-1.3B \cite{wan2025wan}. The model encodes 77 context frames and temporally downsamples them by a factor of 4 via a 3D VAE. For our proposed modules, the memory tokenizer employs a 3D convolution with a kernel size of $2 \times 4\times4$. In the Dynamic Retrieval Attention, the retrieval token length is set to 10, and the local window for the denoising latent is 5. We train our model on the proposed HM-World dataset for 10K iterations using 32 GPUs, with a total batch size of 32.

\noindent\textbf{Evaluation Protocol}. To evaluate our method, we construct a test set comprising 1000 video samples randomly selected from the HM-World dataset, including scenes and subjects that are unseen during training to assess generalization. Our evaluation metrics span three categories: 1) \textbf{General Memory Capacity}. PSNR, SSIM, and LPIPS analyze pixel-wise differences across frames to measure overall reconstruction fidelity. 2) \textbf{Frame-level Consistency}. We adopt Subject Consistency and Background Consistency from the Vbench \cite{huang2024vbench} to measure frame-level coherence. 3) \textbf{Dynamic Subject Consistency (DSC)}. To isolate and evaluate the motion and appearance consistency of moving subjects, especially in re-entering events. We propose a new metric $\textbf{DSC}$ (\textbf{D}ynamic \textbf{S}ubject \textbf{C}onsistency). Specifically, we utilize bounding boxes of moving subjects, which are obtained via YOLOv11\cite{khanam2024yolov11}, to crop the subject regions from the predicted video, the GT video, and the context video. We then extract semantic features from these cropped regions using a pretrained CLIP\cite{radford2021learning} model. After spatial alignment and temporal normalization, we calculate the feature similarities to yield two scores $\text{DSC}_{ctx}$ and $\text{DSC}_{GT}$, formulated as:
\begin{equation}
    \text{DSC}_{GT} = \text{sim}\big( F^{pred}, F^{gt} \big), \quad \text{DSC}_{ctx} = \text{sim}\big( F^{pred}, F^{ctx} \big),
\end{equation}
where \( \text{sim}(\cdot, \cdot) \) refers to the spatially averaged cosine similarity across the feature channels, \( F^{pred} \), \( F^{gt} \), and \( F^{ctx} \) denote subject features from predicted video, GT video, and context video. \( \text{DSC}_{GT} \) evaluates motion and appearance fidelity against the ground truth, while \( \text{DSC}_{ctx} \) evaluates against historical context. 

\begin{table}[t!]
\centering
\scriptsize
\renewcommand{\tabcolsep}{0.6mm} 
\caption{Quantitative comparison with other methods.}
\label{tab:main_table}
\vspace{-10pt}
\begin{tabular}{@{}llccccccc@{}} 
\toprule
Method & Reference & PSNR & SSIM & LPIPS & $\text{DSC}_{ctx}$ & $\text{DSC}_{GT}$ & \begin{tabular}[c]{@{}c@{}}Subj.\\Cons.\end{tabular} & \begin{tabular}[c]{@{}c@{}}Bg.\\Cons.\end{tabular} \\
\midrule
Baseline & - & 18.696 & 0.517 & 0.356 & 0.812 & 0.837 & 0.903 & 0.925\\
DFoT\cite{song2025history} & ICML 25 & 17.693 & 0.482 & 0.410 & 0.803 & 0.826 & 0.893 & 0.913\\
Context-as-Memory\cite{yu2025context} & SIGGRAPH Asia 25 & 18.921 & 0.530 & 0.342 & 0.816 & 0.839 & 0.911 & 0.922 \\
\midrule
\textbf{HyDRA (ours)} & - & \textbf{20.357} & \textbf{0.606} & \textbf{0.289} & \textbf{0.827} & \textbf{0.849} & \textbf{0.926} & \textbf{0.932}\\
\bottomrule
\end{tabular}
\vspace{-10pt}
\end{table}

\subsection{Main Results}

\begin{figure}[t]
    \centering 
    \includegraphics[width=\linewidth]{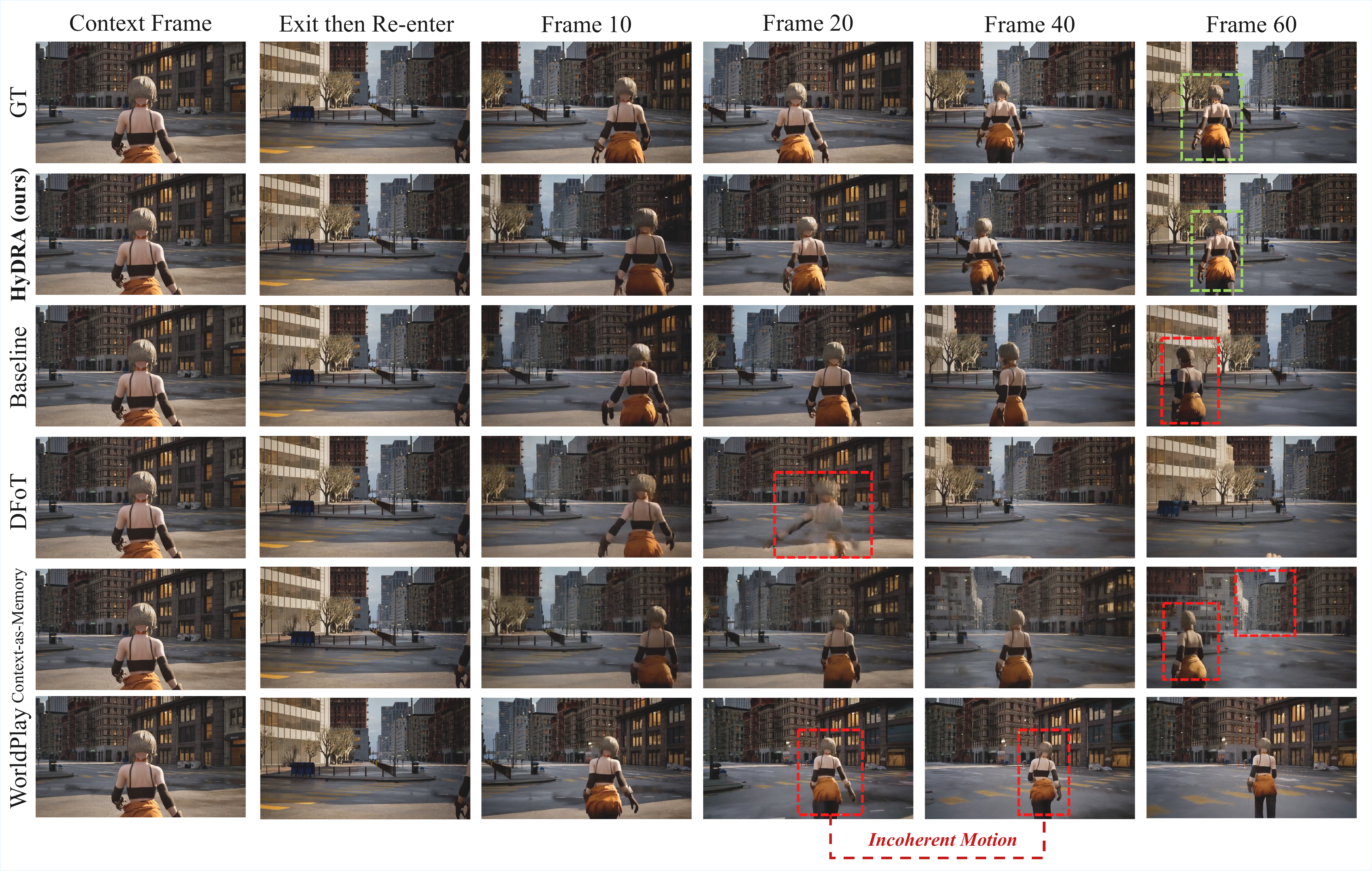}
    \vspace{-20pt} 
    \caption{Qualitative comparison with other methods. The green boxes in the figure represent consistently generated subjects, while the red boxes stand for failure cases.}
    \label{fig:qulitative}
    \vspace{-10pt} 
\end{figure}

In this section, we evaluate the performance of our proposed method against a baseline and state-of-the-art approaches, including DFoT\cite{song2025history} and Context-as-Memory\cite{yu2025context}. The baseline is built upon a  Wan2.1-T2V-1.3B model equipped with a camera encoder, which directly concatenates the context latents and the noisy latents as the input of the DiT.  For fair comparisons, these models are trained on our dataset, strictly adhering to the same training configurations used for our approach. Furthermore, we include a zero-shot evaluation of WorldPlay\cite{sun2025worldplay}, a cutting-edge commercial known for its exceptional consistency. The comparison results are summarized in Tab.~\ref{tab:main_table}, Tab.~\ref{tab:sota-compare} and Fig.~\ref{fig:qulitative}. 

\begin{table}[t!]
\scriptsize
\setlength{\tabcolsep}{1.7mm}
\centering
\caption{Quantitative comparison against the state-of-the-art commercial model.}
\label{tab:sota-compare}
\vspace{-10pt}
\begin{tabular}{ l l c c c c c c c }
 \toprule
 Method &  PSNR & SSIM & LPIPS & \begin{tabular}[c]{@{}c@{}}$\text{DSC}_{ctx}$\end{tabular} & \begin{tabular}[c]{@{}c@{}}$\text{DSC}_{GT}$\end{tabular} & \begin{tabular}[c]{@{}c@{}}Subject\\ Consistency\end{tabular} & \begin{tabular}[c]{@{}c@{}}Background\\ Consistency\end{tabular} \\
 \midrule
WorldPlay\cite{sun2025worldplay}  & 14.855 &  0.355 &  0.500 & 0.822 &  0.832 & 0.910 & 0.925\\
   \textbf{HyDRA (ours)}&   \textbf{20.357} & \textbf{0.606} & \textbf{0.289} & \textbf{0.827} & \textbf{0.849} & \textbf{0.926} &  \textbf{0.932}\\
  % &  &  &  &  &  &  &  \\
\bottomrule
\end{tabular}
\vspace{-15pt}
\end{table}

\textbf{Quantitative Comparison}. As shown in Tab.~\ref{tab:main_table}, HyDRA consistently outperforms competing approaches across all evaluation metrics. Compared to the baseline, our model achieves significant improvements, lifting PSNR from 18.696 to 20.357 and SSIM from 0.517 to 0.606. This demonstrates that HyDRA achieves superior reconstruction accuracy for future frames. Crucially, our method attains the highest $\text{DSC}_{ctx}$ and $\text{DSC}_{GT}$ scores of 0.827 and 0.849, respectively, proving its robust capability to track subjects and maintain their appearance and motion consistency, both in aligning with historical context and predicting future states. The Subject Consistency of 0.926 and Background Consistency of 0.932 further corroborate that it successfully anchors the static stage while preserving overall visual coherence.  While DFoT relies on a neighbor context window, yielding a PSNR of 17.693, and Context-as-Memory utilizes FOV-based context filtering, yielding 18.921, our method surpasses them both, likely because we leverage retrieval over richer token representations and fuse spatiotemporal relationships via dynamic retrieval attention. Tab.~\ref{tab:sota-compare} presents the comparison with the zero-shot performance of WorldPlay. Our method surpasses WorldPlay across all metrics, with a notable PSNR gap of 5.502. Although WorldPlay exhibits lower performance on GT-referenced metrics (e.g., PSNR of 14.855, $\text{DSC}_{GT}$ of 0.832) due to domain distribution gap and lack of specific finetuning, it demonstrates remarkable robustness on context-referenced metrics by achieving a $\text{DSC}_{ctx}$ of 0.822. This observation not only confirms that extensively trained models possess fair hybrid consistency but also indirectly validates the rationality of our proposed DSC metrics in reflecting dynamic subject consistency. Ultimately, these impressive results highlight the exceptional capabilities of our model, demonstrating its superiority even over established commercial models.

\textbf{Qualitative Comparison}. We present a qualitative comparison in Fig.~\ref{fig:qulitative}. In the case of complex exit-and-entry events, the baseline and Context-as-Memory exhibit severe subject distortion and motion incoherence. DFoT fails to maintain subject integrity, leading to complete vanishing. While WorldPlay manages to preserve the subject's appearance consistency, it suffers from stuttering movements and unnatural actions. In contrast, our method successfully maintains hybrid consistency, preserving both the subject's identity and motion coherence after the subject re-enters the frame. Due to space limitations, more generation results are provided in the \textbf{\underline{supplementary materials}}.

\vspace{-10pt}
\subsection{Ablation Study}
In this section, we conduct comprehensive ablation studies to validate the effectiveness of the core components in our method.
% \vspace{-10pt}

% \vspace{-10pt}

\begin{table}[t]
\scriptsize
\setlength{\tabcolsep}{2.1mm}
\centering
\caption{Kernel Size of Memory Tokenizer.}
\label{tab:tokenizer}
\vspace{-10pt}
\begin{tabular}{ c c c c c c c c c }
 \toprule
 $T$ & $H\times W$ & PSNR & SSIM & LPIPS & \begin{tabular}[c]{@{}c@{}}$\text{DSC}_{ctx}$\end{tabular} & \begin{tabular}[c]{@{}c@{}}$\text{DSC}_{GT}$\end{tabular} & \begin{tabular}[c]{@{}c@{}}Subject\\ Consistency\end{tabular} & \begin{tabular}[c]{@{}c@{}}Background\\ Consistency\end{tabular} \\
 \midrule
 $2$ & $2\times2$& 20.113 &  0.599& 0.299 & 0.820 & 0.843 &  0.919& 0.929 \\
  $2$ &$4\times4$& \textbf{20.357} & 0.606 & \textbf{0.289} & \textbf{0.827} & \textbf{0.849} & \textbf{0.926} &  \textbf{0.932}\\
  $2$ & $8\times8$& 20.230 & \textbf{0.610} & 0.292 & 0.822 & 0.843 & 0.923 & 0.927 \\
  $1$ & $4\times4$&  19.076&  0.554& 0.337 & 0.819 & 0.841 & 0.912 & 0.925 \\
  % &  &  &  &  &  &  &  \\
\bottomrule
\end{tabular}
\vspace{-10pt}
\end{table}
% \vspace{-10pt}
\textbf{Kernel Size of Memory Tokenizer}. We first evaluate the impact of different kernel sizes in the memory tokenizer, with the results summarized in Tab.~\ref{tab:tokenizer}. The kernel size is denoted as $T\times H\times W$, representing the temporal, height, and width dimensions, respectively. The results indicate that our model exhibits strong robustness to variations in the spatial dimensions. The performance differences among spatial dimensions' settings are marginal, as transitioning from the optimal $4\times4$ configuration to $2\times2$ or $8\times8$ results in a minor PSNR decrease of only 0.244 and 0.127, respectively. In contrast, when the temporal dimension is reduced to 1, we observe a significant performance drop of 1.281 in PSNR and 0.014 in $\text{DSC}_{GT}$, which demonstrates the necessity of temporal interaction within the tokenizer for capturing long-term dynamic information.

% \vspace{-15pt}

% \vspace{-5pt}
\textbf{Number of Retrieved Tokens}. We investigate the effect of the retrieved memory token length in Tab.~\ref{tab:length}. Retrieving only 5 tokens yields suboptimal performance with a PSNR of 19.309, indicating that an overly restricted token count leads to severe information loss. Conversely, increasing the number to 10 and 15 generates better and more stable results, with negligible differences between the two. This suggests that a moderate number of tokens is sufficient to provide the necessary spatiotemporal information without introducing redundant noise.
% \vspace{-15pt}

\begin{table}[t]
\scriptsize
\setlength{\tabcolsep}{2.5mm}
\centering
\caption{Number of retrieved tokens.}
\label{tab:length}
\vspace{-10pt}
\begin{tabular}{ c c c c c c c c }
 \toprule
 Setting & PSNR & SSIM & LPIPS & \begin{tabular}[c]{@{}c@{}}$\text{DSC}_{ctx}$\end{tabular} & \begin{tabular}[c]{@{}c@{}}$\text{DSC}_{GT}$\end{tabular} & \begin{tabular}[c]{@{}c@{}}Subject\\ Consistency\end{tabular} & \begin{tabular}[c]{@{}c@{}}Background\\ Consistency\end{tabular} \\
 \midrule
 5& 19.309 &  0.566& 0.339 & 0.817 & 0.836 & 0.913 & 0.927 \\
  10&   \textbf{20.357} & 0.606 & \textbf{0.289} & 0.827 & \textbf{0.849} & \textbf{0.926} &  0.932\\
  15& 20.333 & \textbf{0.612} & 0.291 &  \textbf{0.828}&  0.842& 0.925 & \textbf{0.935} \\
  % &  &  &  &  &  &  &  \\
\bottomrule
\end{tabular}
% \vspace{-15pt}
\end{table}
% \vspace{-10pt}
\textbf{Token Retrieval Approaches}. We ablate the token retrieval mechanism by comparing our dynamic affinity retrieval with FOV overlap retrieval in Tab.~\ref{tab:retrieve_way}. Since a single memory token in our architecture aggregates information from multiple frames with varying camera poses, we average the camera poses of the source frames to represent the token's pose. We then follow Context-as-Memory \cite{yu2025context} to calculate the FOV overlap between the token and the target frame to perform retrieval. Experimental results demonstrate that our method outperforms the FOV-based approach across all metrics, notably improving Subject Consistency from 0.908 to 0.926. This superiority stems from leveraging QK interactions to assess fine-grained spatiotemporal relevance, whereas the FOV-based approach relies solely on static geometry overlap.

% \vspace{-15pt}

\begin{table}[t]
\scriptsize
\setlength{\tabcolsep}{1.8mm}
\centering
\caption{Approaches to retrieve tokens.}
\label{tab:retrieve_way}
\vspace{-10pt}
\begin{tabular}{ c c c c c c c c }
 \toprule
 Method & PSNR & SSIM & LPIPS & \begin{tabular}[c]{@{}c@{}}$\text{DSC}_{ctx}$\end{tabular} & \begin{tabular}[c]{@{}c@{}}$\text{DSC}_{GT}$\end{tabular} & \begin{tabular}[c]{@{}c@{}}Subject\\ Consistency\end{tabular} & \begin{tabular}[c]{@{}c@{}}Background\\ Consistency\end{tabular} \\
 \midrule
 FOV Overlap& 19.776 &  0.586&  0.300& 0.820 &  0.844&  0.908&  0.930\\
 Dynamic Affinity& \textbf{20.357} & \textbf{0.606} & \textbf{0.289} & \textbf{0.827} & \textbf{0.849} & \textbf{0.926} &  \textbf{0.932}\\ 

  % &  &  &  &  &  &  &  \\
\bottomrule
\end{tabular}
\vspace{-10pt}
\end{table}

\section{Conclusion}
In this paper, we introduce the novel paradigm of \textbf{Hybrid Memory}, challenging models to simultaneously maintain static background consistency and dynamic subject coherence, particularly during complex exit-and-re-entry events. To systematically facilitate research in this field, we construct \textbf{HM-World}, the first large-scale video dataset dedicated to hybrid memory, featuring highly diverse scenarios and complex dynamic processes. To tackle the challenge of hybrid memory, we propose \textbf{HyDRA}, an advanced memory architecture specifically designed to effectively extract and retrieve motion and appearance cues for consistent generation. Extensive experiments demonstrate that HyDRA significantly outperforms existing methods. We hope that the hybrid memory paradigm, alongside the HM-World dataset and the HyDRA framework, will inspire new research and provide a solid foundation for advancing video world models.

\textbf{Limitations and Future Work}. Despite the promising results, our work still presents certain limitations. Specifically, HyDRA's performance in maintaining consistent generation tends to degrade in highly complex scenes involving three or more subjects or severe occlusions. In future work, we plan to explore more advanced and robust memory mechanisms to handle intricate multi-subject dynamics and scale our approach to unconstrained real-world environments.

\section*{Acknowledgements} 
We express our sincere gratitude to Jichao Wang, Xiaole Xiong, Siyuan Luo, Mengyuan Li, Boyu Zheng, and Yike Yin from Kuaishou Technology for their invaluable assistance in developing the HM-World dataset.

\bibliographystyle{splncs04}
\bibliography{main}

\clearpage
\appendix 

\title{Out of Sight but Not Out of Mind: Hybrid Memory for Dynamic Video World Models} 
\titlerunning{Hybrid Memory for Dynamic Video World Models}
\author{} 
\institute{}
\date{} 
\maketitle

\vspace{-30pt}
\begin{center}
    \Large Supplementary Material
    % \vspace{20pt}
\end{center}

This file provides additional information about our work, mainly from more generation results and ablation studies.

\vspace{-5pt}
\begin{center}
    \includegraphics[width=\linewidth]{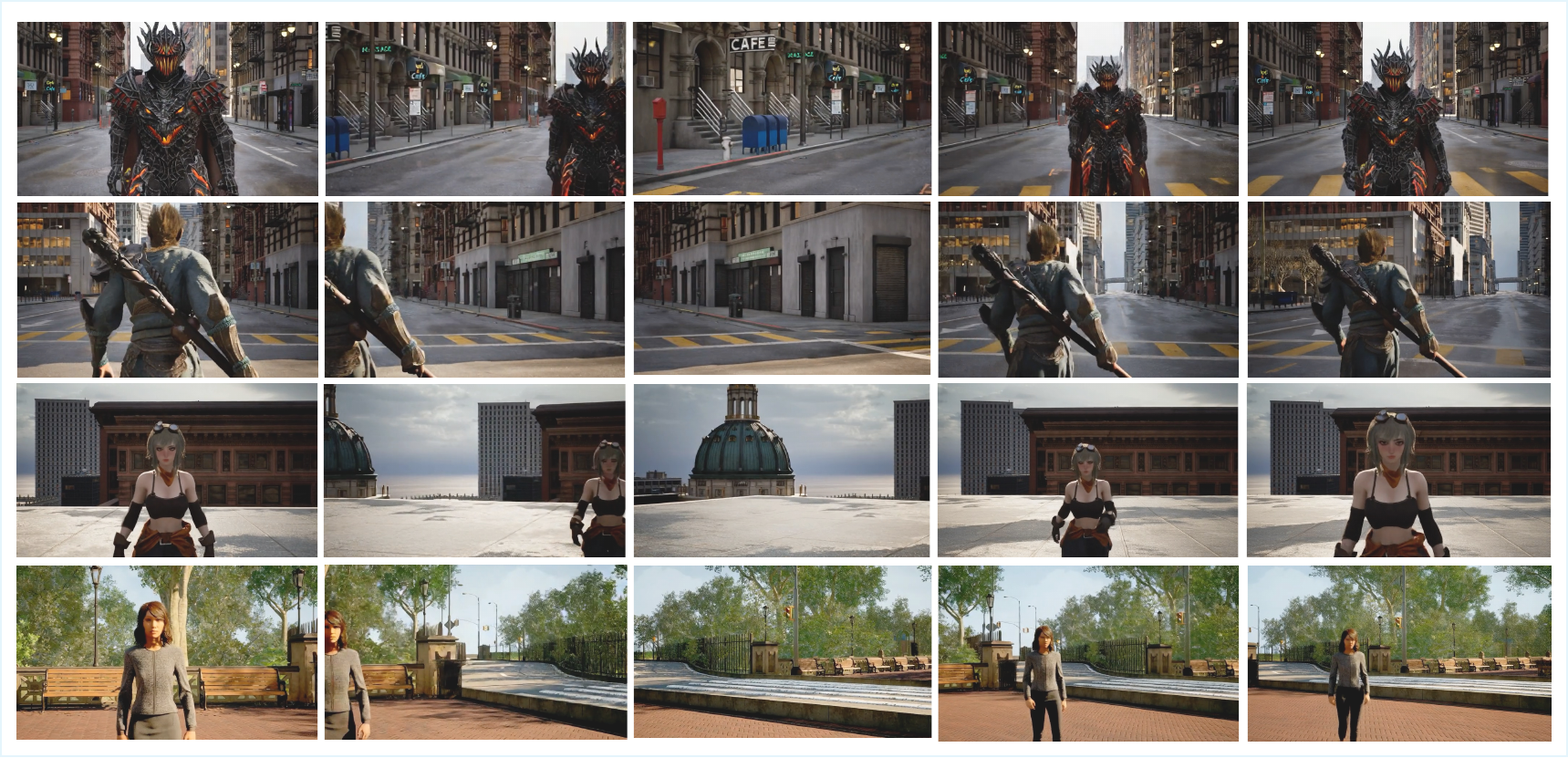}
    \vspace{-17pt} % 根据需要调整图注和图片的间距
    \captionof{figure}{The results  generated by HyDRA.}
    \label{fig:generation_results}
\end{center}
\vspace{-20pt}

\section{Qualitative Analysis}

\subsection{Generation Results}
Fig.~\ref{fig:generation_results} shows HyDRA's generation results across multiple scenes, subjects, and trajectories. HyDRA effectively implements memorization of both background and subjects in complex dynamic scenarios with exit-entry events, maintaining appearance and motion consistency.

\subsection{Open-Domain Results}
\begin{figure}[t]
	\begin{center}
		\includegraphics[width=\linewidth]{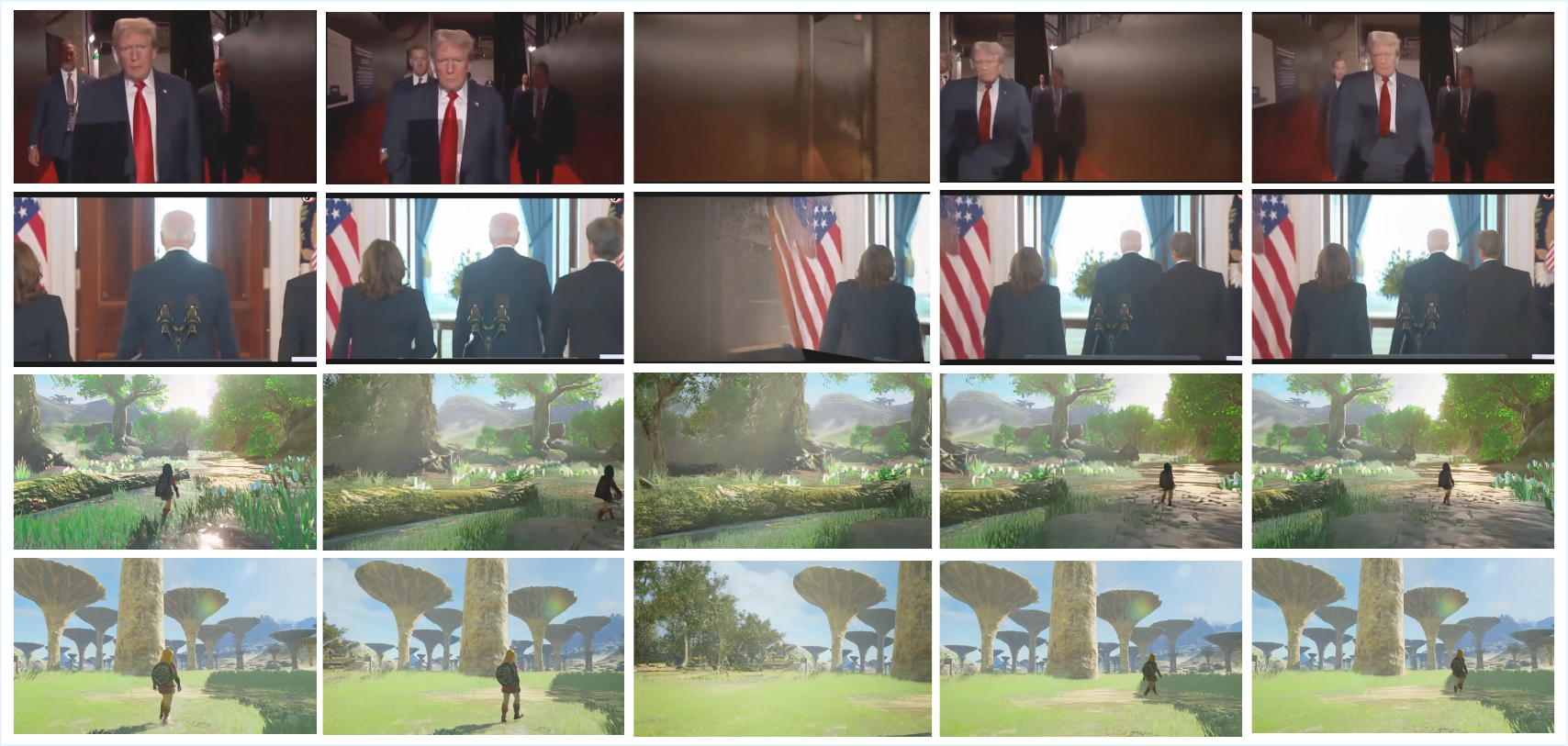}
	\end{center}
	\vspace{-15pt}
	\caption{Open-domain results of HyDRA.
 }
 \vspace{-5pt}
	\label{fig:opendomain}
\end{figure}

We collect open-domain videos featuring subject motion from the Internet and apply back‑and‑forth camera movements for inference. The results in Fig~\ref{fig:opendomain} demonstrate that even in entirely unseen scenes, HyDRA exhibits good capacity of hybrid memory.

% This section 

\section{More Ablation Studies}

In this section, we further conduct comprehensive ablation analyses on our proposed method and core designs.

\begin{figure}[t]
	\begin{center}
		\includegraphics[width=\linewidth]{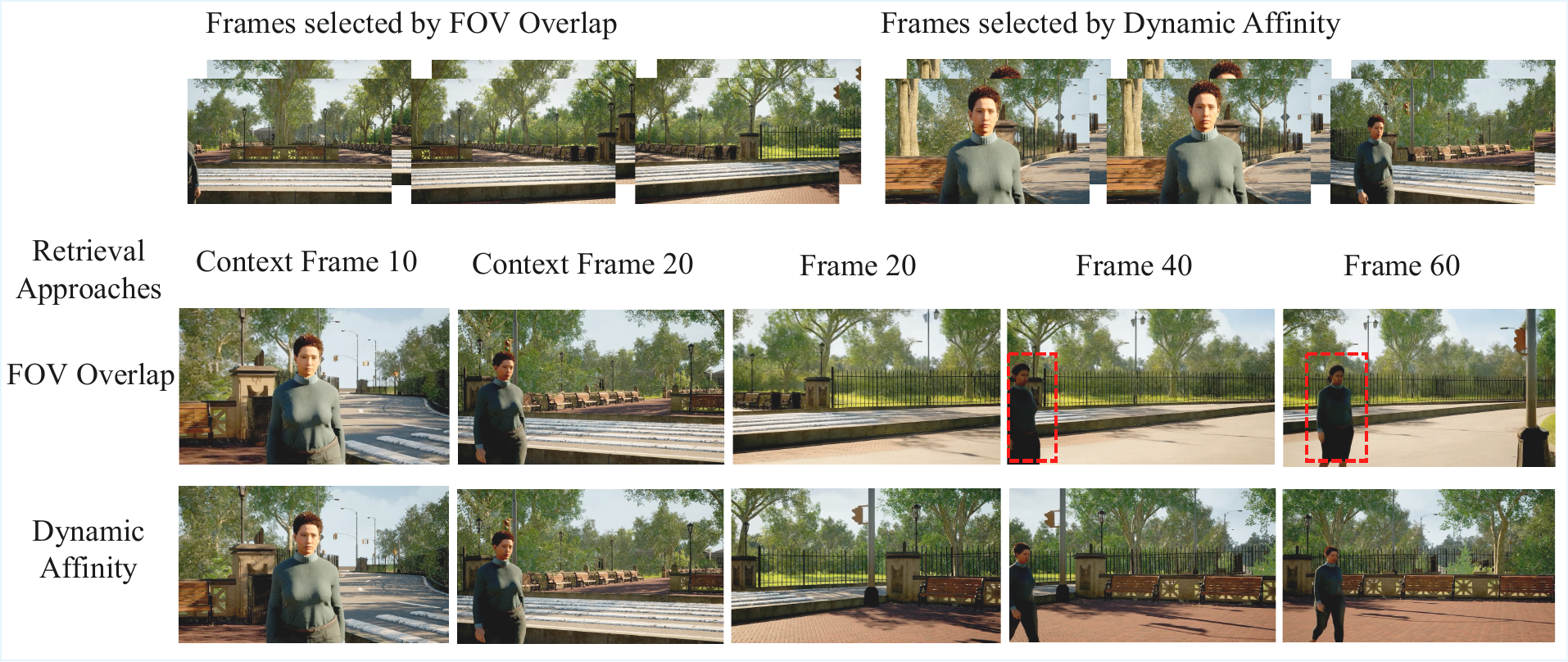}
	\end{center}
	\vspace{-15pt}
	\caption{Qualitative comparison between retrieval methods. The upper displays frames selected by different methods, while the lower shows the generation results. Selected frames are the source frames of the selected tokens.
 }
 \vspace{-15pt}
	\label{fig:abla_retrieve}
\end{figure}

\begin{figure}[t]
	\begin{center}
		\includegraphics[width=\linewidth]{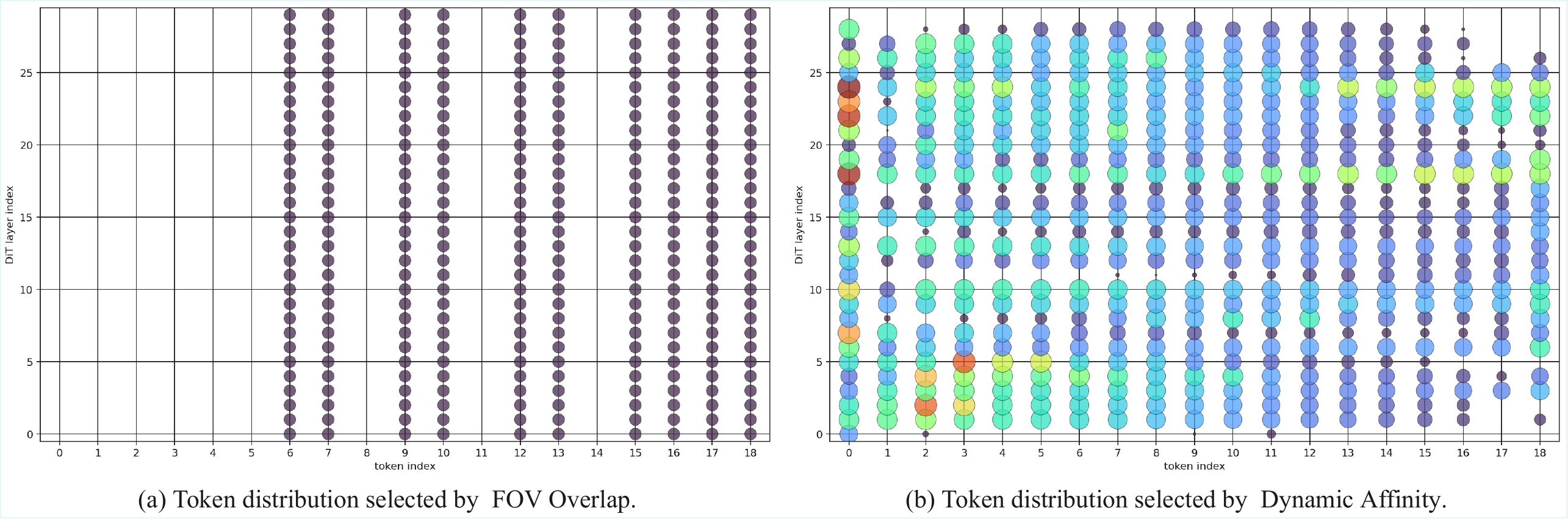}
	\end{center}
	\vspace{-15pt}
	\caption{Distribution comparison of different retrieval methods. The x-axis and y-axis represent the token index and DiT layers, respectively. The bubble size and color reflect the selection frequency of each token during the entire denoising process. (a) The FOV overlap method yields a fixed token selection. (b) Our dynamic affinity method exhibits a diverse retrieval distribution, enabling the perception of richer memory contexts.
 }
 \vspace{-10pt}
	\label{fig:retrieve_analyze}
\end{figure}

\textbf{Analysis of Retrieval Approaches}. We first compare our dynamic-affinity-based retrieval method with the traditional Field of View (FOV) overlap filtering approach. As illustrated in Fig.~\ref{fig:abla_retrieve}, during a long camera movement involving complex exit-and-re-entry events, the FOV-based method merely selects the nearest camera poses corresponding to the re-entry clip. Consequently, it mistakenly retrieves empty shots, leading to a severe loss of critical appearance information and inconsistent generation. In contrast, our dynamic affinity approach filters memory tokens based on feature-level correlations. It successfully retrieves keyframes containing rich subject details, thereby maintaining the appearance and motion consistency of the subject after re-entry. Furthermore, we investigate the distribution of the retrieved tokens across different filtering strategies in Fig.~\ref{fig:retrieve_analyze}. The FOV overlap method relies on static 3D geometric calculations, meaning the selected memory tokens remain fixed throughout the entire inference stage. In contrast, our dynamic affinity method computes feature-level correlations dynamically. As a result, it adaptively selects different tokens at different timesteps and across different DiT layers. This dynamic mechanism grants the model a broader memory receptive field and superior flexibility during the generation process.

\begin{figure}[t]
	\begin{center}
		\includegraphics[width=\linewidth]{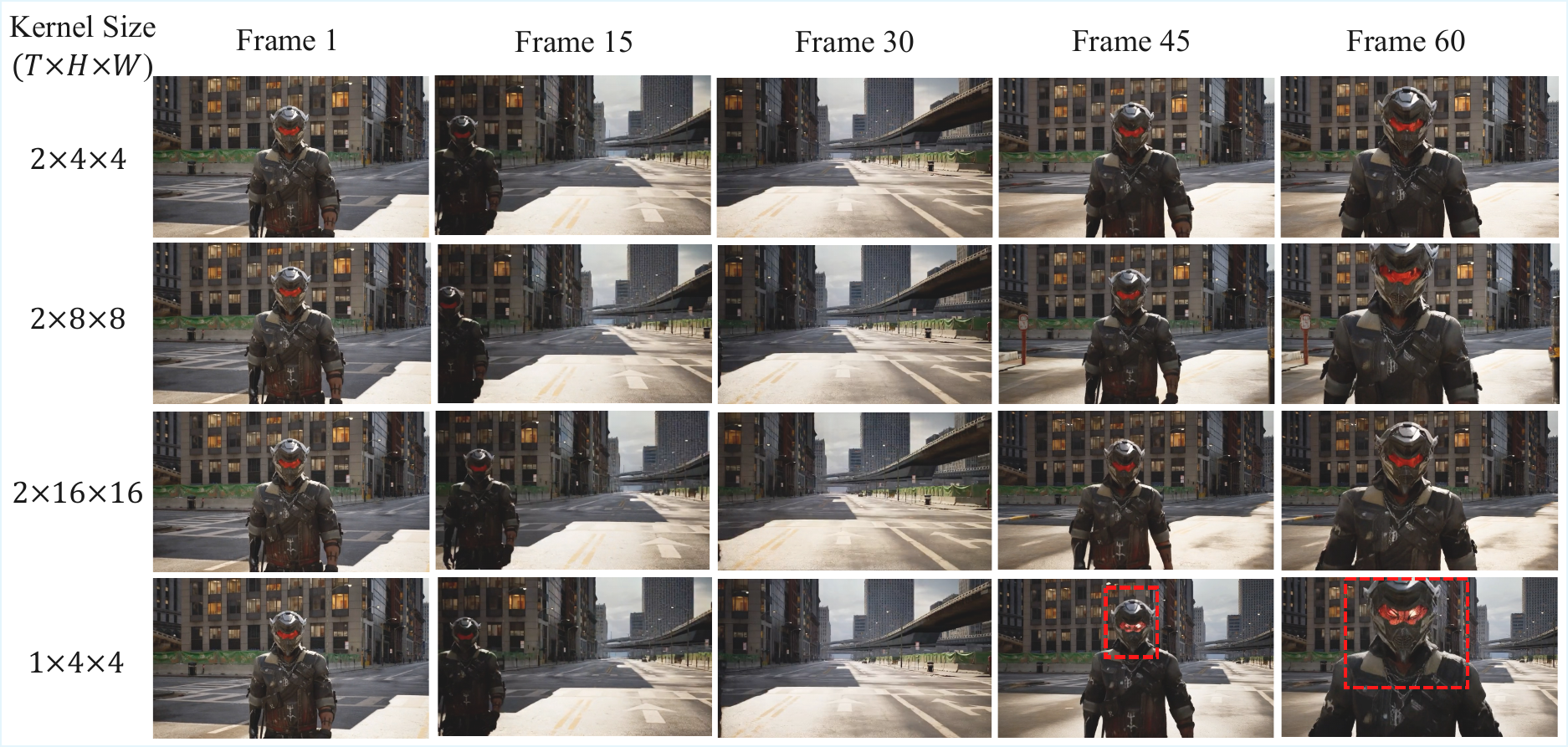}
	\end{center}
	\vspace{-15pt}
	\caption{Qualitative comparison between different kernel sizes of the Memory Tokenizer. The red bounding boxes annotate the inconsistent region.
 }
 \vspace{-15pt}
	\label{fig:abla_kernel}
\end{figure}

\textbf{Ablation on Kernel Size of Memory Tokenizer}. We provide further qualitative ablation results regarding the kernel size of the Memory Tokenizer. As shown in Fig.~\ref{fig:abla_kernel}, when the temporal dimension of the kernel size is set to $2$, the generated results maintain spatiotemporal consistency due to effective temporal interaction. However, when the temporal kernel size is reduced to $1$ (i.e., no temporal interaction during tokenization), noticeable inconsistencies emerge in the generated subjects. These qualitative observations further corroborate the quantitative ablation results presented in the main paper.

\begin{figure}[t!] 
	\begin{center}
		\includegraphics[width=\linewidth]{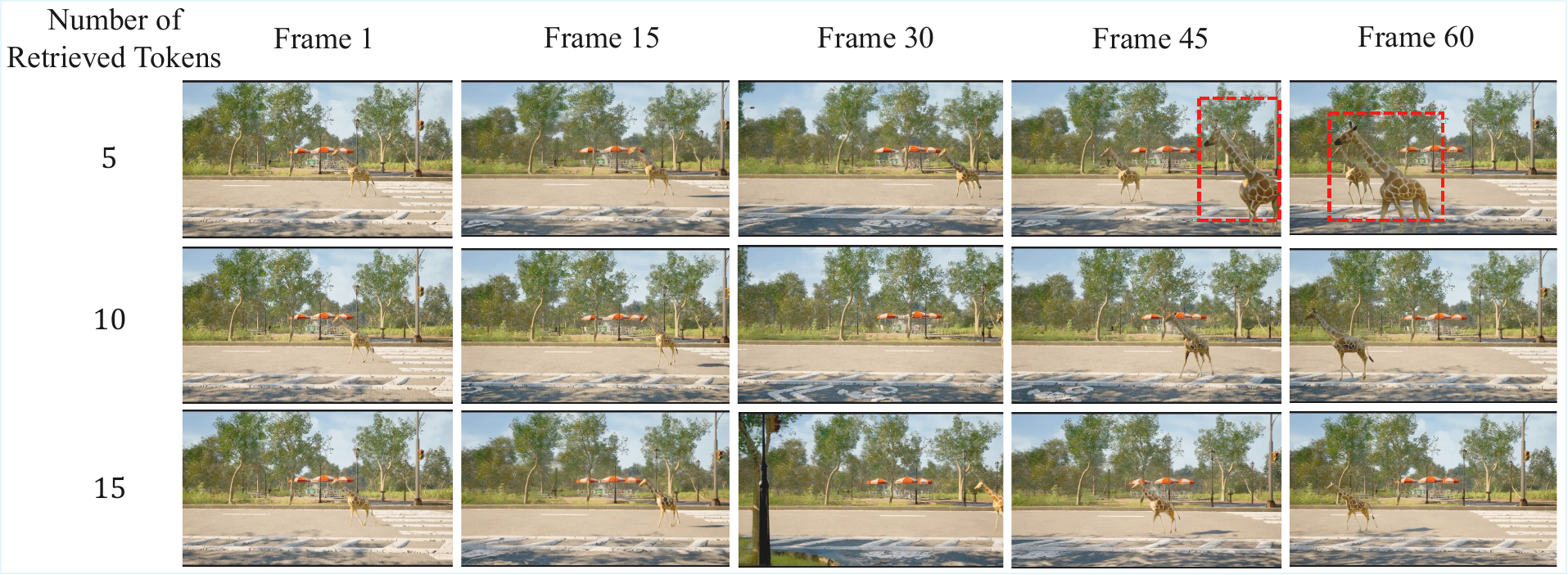}
	\end{center}
	\vspace{-15pt}
	\caption{Qualitative comparison between the number of retrieved tokens. The red bounding boxes annotate the inconsistent region.}
	\label{fig:abla_length}
	
	\vspace{-5pt} 
	
	\begin{center}
		\includegraphics[width=\linewidth]{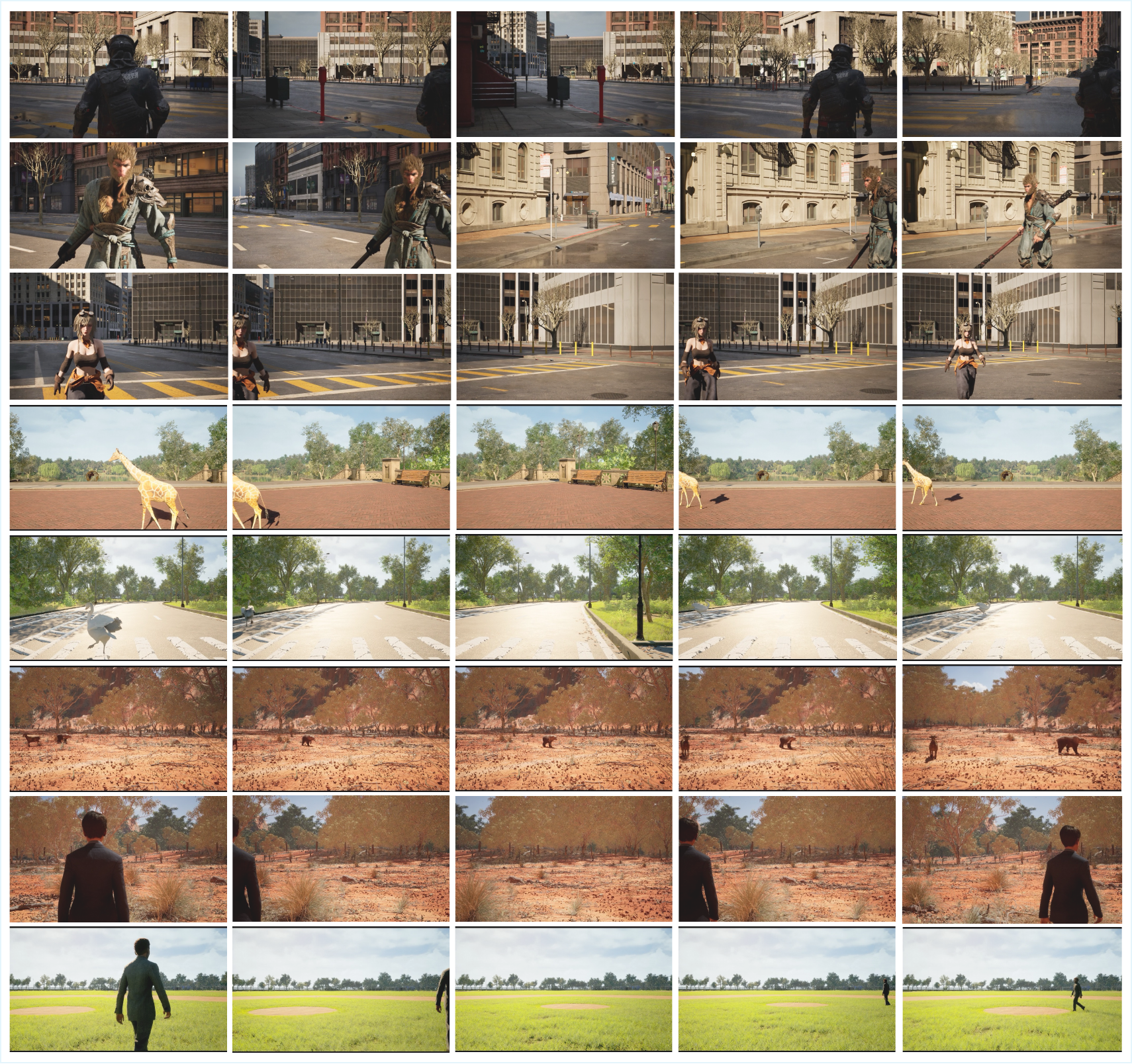}
	\end{center}
	\vspace{-15pt}
	\caption{Additional examples of HM-World dataset.}
    \vspace{-15pt}
	\label{fig:dataset_vis}
\end{figure}

\textbf{Ablation on Number of Retrieved Tokens}. We qualitatively ablate the number of retrieved tokens. As depicted in Fig.~\ref{fig:abla_length}, restricting the token length to $5$ results in a substantial loss of context information, which misleads the model into generating severe artifacts (e.g., hallucinating two giraffes instead of one). In contrast, other settings with an adequate number of retrieved tokens successfully maintain subject consistency and physical plausibility.

\section{Additional Examples from the HM-World Dataset}

To further illustrate the challenges present in the proposed HM-World dataset, we provide additional examples in Fig.~\ref{fig:dataset_vis}.

\end{document}